\documentclass{llncs}
\usepackage{makeidx}  
\usepackage{amsfonts}
\usepackage{amsmath}
\usepackage{graphicx}
\usepackage{subcaption}

\begin{document}
\frontmatter          
\pagestyle{headings}  
%
\mainmatter              
\title{TridentNet: A Conditional Generative Model for Dynamic Trajectory Generation}
\titlerunning{TridentNet}  
%
\author{David Paz, Hengyuan Zhang, \and Henrik I.~Christensen}
\authorrunning{Paz D.~et~al.} 
%
%
\institute{University of California, San Diego, La Jolla, CA 92093, USA\\
\email{dpazruiz, hyzhang, hichristensen@ucsd.edu}
}

\maketitle              

\begin{abstract}
In recent years, various state of the art autonomous vehicle systems and architectures have been introduced. These methods include planners that depend on high-definition (HD) maps and models that learn an autonomous agent's controls in an end-to-end fashion. While end-to-end models are geared towards solving the scalability constraints from HD maps, they do not generalize for different vehicles and sensor configurations. To address these shortcomings, we introduce an approach that leverages lightweight map representations, explicitly enforcing geometric constraints, and learns feasible trajectories using a conditional generative model. Additional contributions include a new dataset that is used to verify our proposed models quantitatively. The results indicate low relative errors that can potentially translate to traversable trajectories. The dataset created as part of this work has been made available online.\footnote{Dataset: avl.ucsd.edu} 

\end{abstract}

\section{Introduction}
\label{sec:intro}
Autonomous vehicle technology has been developing at a fast rate in recent years. Key issues to consider in the design of autonomous vehicles are: where am I (localization)? where am I going (planning)? and how do I get there (control)? Localization has mainly been addressed using HD maps. For planning, reliance on HD maps has also been dominant, but most importantly dynamic planning in the presence of other road-users is still a major challenge, whereas control/trajectory following is largely a solved problem. 

\begin{figure*}
    \centering
    \includegraphics[scale=0.225]{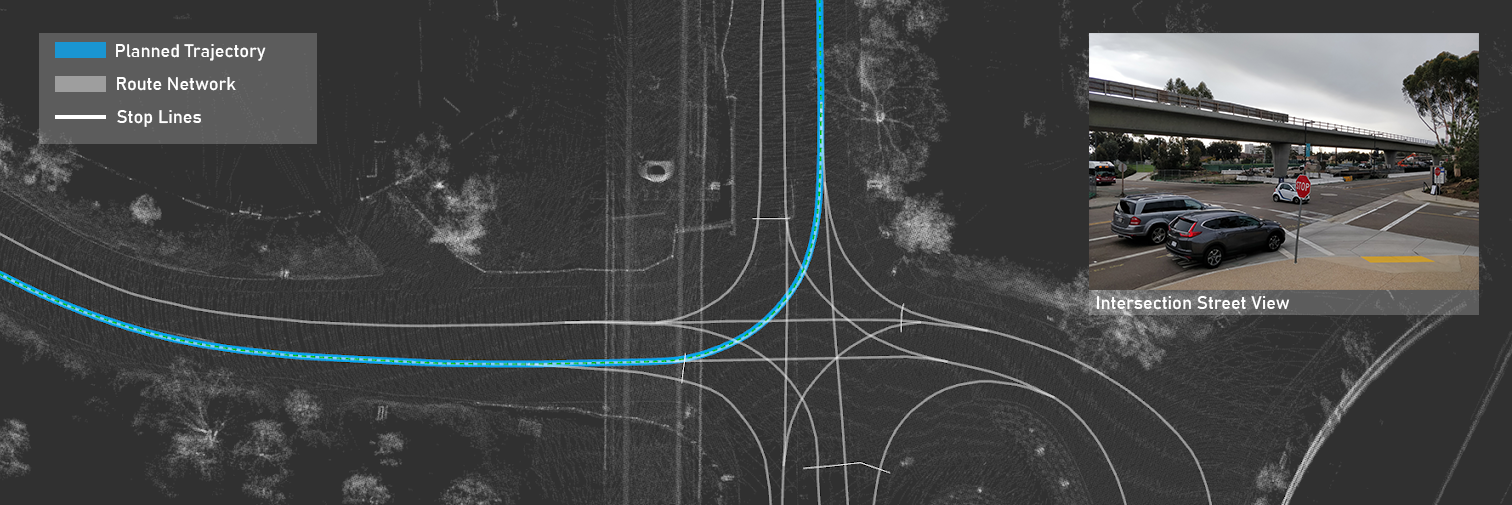}
    \caption{A simplified HD map definition for a UC San Diego four-way intersection. The route network, the planned trajectory and stop-line definitions are coded in gray, light-blue and white line segments, respectively. An image from a bystander perspective is shown on the top-right.}
    \label{fig:hd-map}
\end{figure*}

A big challenge, both for localization and planning, is scalability due to their dependencies on HD maps. 
These HD maps used in planning present a bottleneck for large scale applications as the environments drastically grow, and they require extensive maintenance and manual labeling. They generally include contextual information such as the locations and dimensions of crosswalks, lane markings, stop signs, traffic signs and even center-line definitions. A number of open source datasets have been introduced with similar labels \cite{nuscenes,argoverse}. Fig. \ref{fig:hd-map} shows a simplified version of an HD map from a previous work \cite{paz:hd-maps} that includes only centerlines and stop lines. Despite the simplification, annotating the features with centimeter-level details and maintaining them during road construction or highly dynamic environments require a considerable amount of effort and is not scalable.

For these reasons, end-to-end models have been introduced to learn the control actions required to perform a maneuver without detailed map representations. For example, \cite{hecker:one} and \cite{rus:one} use coarse map representations and raw camera data to learn a control action for the next time step. One advantage of this approach is that the models operate with lightweight map representations. Similarly, \cite{hecker:two} uses a Generative Adversarial Network(GAN) approach; however, the proprietary map representations may require considerable manual labeling. Reinforcement learning methods for full-scale vehicles have also started to be explored \cite{rus:two,capaso:iv}. Nevertheless, these state of the art end-to-end models do not enforce geometric constraints and still exhibit scalability challenges given that the training data needed for the models assumes a fixed vehicle configuration to learn vehicle-specific controls \cite{hecker:one,hecker:two,rus:one,rus:two,capaso:iv}. For this reason, if a model needs to be extended for other vehicles or the sensor calibration is compromised, the performance may be hindered, and new data for training may be required.

To address these challenges, we introduce a conditional generative approach that can dynamically generate ego-centric trajectories using nominal representations. Our model utilizes coarse OpenStreetMap representations to encode the plan required to reach a destination and lightweight local semantic scene models to encode the contextual information around the vehicle such as road markings, drivable areas, and sidewalks. In contrast to existing end-to-end models, our approach can i) generalize to various sensor configurations by manipulating the local semantic scene models using rigid body transformations, and ii) generalize to other vehicle platforms by dynamically generated trajectories instead of vehicle specific control actions.

To summarize, in Section \ref{sec:related} related efforts are presented as a baseline and motivation. In Section~\ref{sec:trident} a dynamic path generation approach, termed TridentNet, is introduced and formulated using a Conditional Variational Autoencoder (CVAE). Section \ref{sec:experiments} presents experiments for validation and the associated public data-setup. To the best of our knowledge, this is the first approach with an accompanying dataset release for dynamic trajectory generation using lightweight representations. For that reason, comparisons with respect to related work are performed from an architecture perspective and the associated implications on scalability. Finally, the paper is summarized in Section \ref{sec:summary} together with a number of challenges for the future.

\section{Related Work}
\label{sec:related}
\subsection{Autonomous Navigation Frameworks}
\label{sec:related-navigation}

Autonomous driving frameworks and architectures that leverage HD maps include Apollo~\cite{apollo} and Autoware~\cite{autoware}. In general, these architectures follow a hierarchical structure with clearly defined interfaces for localization, perception, planning, and control. These layers of abstraction and interfaces allow vehicle specific adaptation by simply extending existing kinematics models and controls. For example, in \cite{paz:fsr}, open-source packages including Autoware and ROS are used as a development architecture for autonomous micro-transit applications using golf carts: an use-case with specific vehicle types that were not explicitly defined for the ecosystem. This can be seen as an advantage in generalizing autonomous vehicle technology to other vehicle types and applications; however, one major limitation involves the maintenance cost of HD maps as the maps grow in size and complexity.

\subsection{Scene Representations}
\label{sec:related-reps}
While HD maps can offer contextual information, various other methods have been proposed for automatically generating scene representations without fully defined road networks. These include probabilistic semantic 2D bird's eye view representations that enforce geometric constraints \cite{paz:iros-semantic} and deep learning based monocular depth estimation techniques \cite{can,roddick} that learn semantic bird's eye view representations from a single monocular camera. These models  have shown promising results when it comes to mitigating the scalability constrains from HD maps; however, unlike HD maps, additional work is required for downstream tasks when it comes to estimating feasible trajectories during planning and navigation. Thus, in this work we focus on leveraging 2D bird's eye view representations to learn feasible ego-centric trajectories without manual labeling. 

\subsection{Trajectory Forecasting}
\label{sec:related-forecast}
An important aspect of our work involves dynamic path generation to be capable of estimating feasible trajectories for the ego-vehicle. Our work is inspired by the latest developments on conditional generative models that have been applied for road user trajectory forecasting including pedestrians and vehicles. While prediction for other road users is out of the scope of this work, we leverage similar methods for trajectory generation. Examples of models utilized for prediction include LSTMs \cite{alahi}, Conditional Variational Autoencoders (CVAEs) \cite{leung,ivanovic}, GANs \cite{gupta}, Graph Convolution Networks (GCNs) \cite{yu}, and most recently with Equivariant Continuous Convolutions (ECCO)~\cite{yu2}.

\section{TridentNet CGM}
\label{sec:trident}
\subsection{Approach}
\label{sec:approach}
In this section, we introduce TridentNet: a conditional generative model for generating dynamic ego-centric trajectories without manually annotated HD maps. Given a global plan $\mathbf{G}_t$ that represents the plan to reach a destination and a local scene representation $\mathbf{L}_t$ at time $t$ that encodes the road features, lane markings and drivable areas, the task at hand becomes estimating a set of traversible waypoints $\mathbf{y} = \mathbf{y}_t, \mathbf{y}_{t+1}, \ldots, \mathbf{y}_{t+H}$ with respect to the ego-vehicle or a map frame-- where $H$ corresponds to the generation horizon. In the next three subsections, we describe the global plan approach and representation ($\mathbf{G}_t$), the local scene representation ($\mathbf{L}_t$), as well as the CVAE approach for modeling the distribution of feasible trajectories $p(\mathbf{y}\mid \mathbf{m})$, where $\mathbf{m}=(\mathbf{G}_t, \mathbf{L}_t)$. An important note to make here is that while obstacle detection, tracking, and motion planning are not covered in this work, the generated trajectory can facilitate the development of interfaces for these tasks. Hence, making it possible to estimate if an obstacle is blocking the road and if evasive maneuvers need to be taken. This is discussed in the conclusion and left for future work.

\begin{figure*}
    \centering
    \includegraphics[scale=0.45]{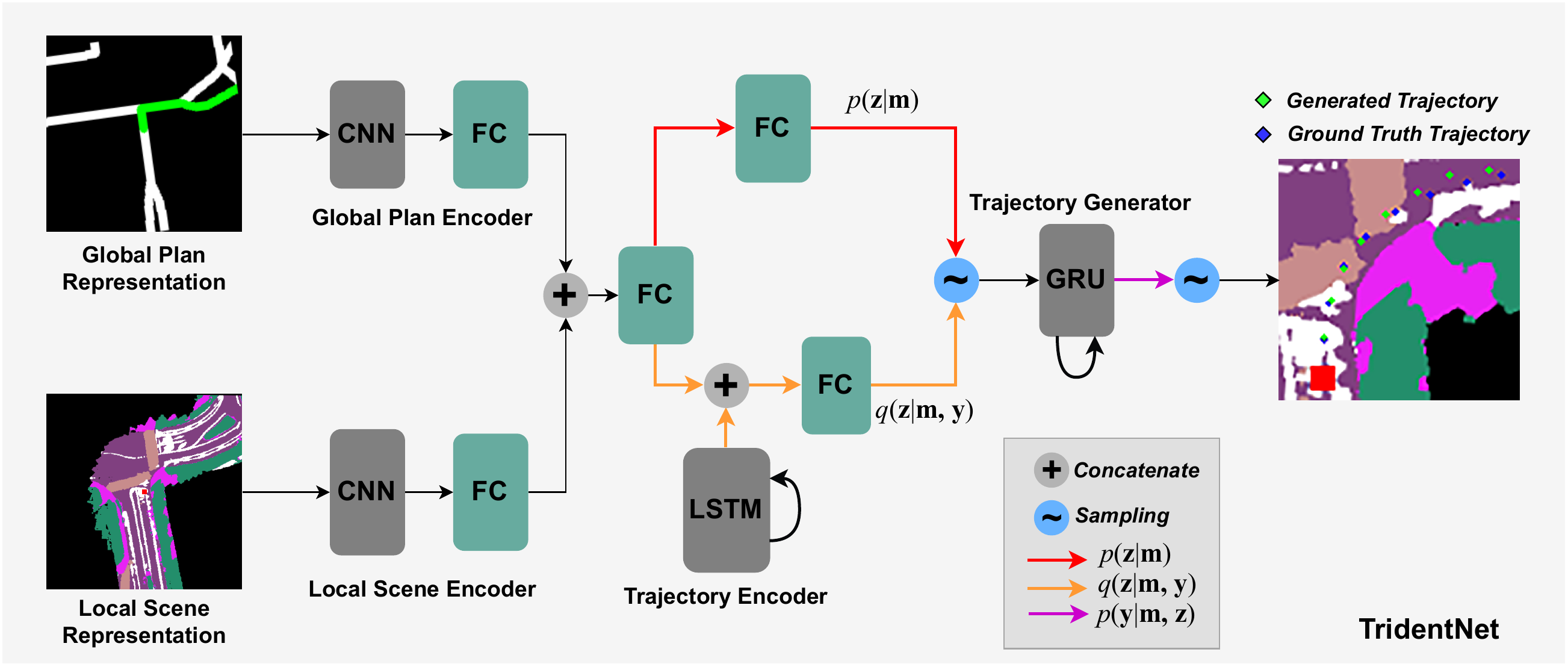}
    \caption{TridentNet leverages lightweight map representations including OpenStreetMaps, local semantic maps and a Conditional Variational Autoencoder to generate trajectories. }
    \label{fig:tridentnet}
\end{figure*}

\subsection{Global Plan}
\label{sec:global-plan}
Navigation requires choosing an optimal maneuver that aligns with a global plan and objective. An example of this involves intersection navigation: when the vehicle is approaching an intersection, there are multiple traversible trajectories but only one should be chosen and executed. This depends on global planning information that describes the actions that need to be taken to reach a destination given the current state of the vehicle. To capture this global plan information and context, we leverage OpenStreetMaps in a similar fashion as \cite{rus:one}. First, we download OpenStreetMaps for the geographical areas of interest, parse its XML format, and build a graph. The graph representation allows us to combine GPS, IMU and odometry information, set a target position, and estimate the shortest path between the position of the autonomous vehicle and the destination using Dijkstra's algorithm. With the path and localization information, we render bird's eye view representations for the global plan as shown on the top left of Fig. \ref{fig:tridentnet}; where the white segments correspond to road segments nearby and the green segment to the trajectory to be taken. 

\subsection{Local Scene Representation}
\label{sec:local-rep}
To estimate trajectories that a self-driving car can execute, sufficient contextual information must be gathered to represent the environment including lane markings, drivable areas and even side walks. To provide these contextual cues, we generate 2D semantic maps. 

Semantic maps can include information about road features that are critical for navigation like roads, lane markings, crosswalks and sidewalks. Our approach for generating semantic maps is based on the models proposed in our prior work \cite{paz:iros-semantic}. By fusing LiDAR and camera data, this approach is capable of automatically generating representations of the world with robust road features in a single pass of the scene without human annotation. 

The algorithm consists of three parts, i) an image based semantic segmentation module, ii) a fusion and association module, and iii) a probabilistic mapping module. The input camera data goes through a semantic segmentation neural network (Deeplab v3+~\cite{deeplab}) with a light-weight backbone to generate labels for each pixel. LiDAR point clouds are then used to generate a dense point cloud map. We leverage localization to crop a portion of the point cloud map that is close to the vehicle. This cropped local dense map is then projected onto the semantic image to associate labels with each point. The point cloud with semantic labels is projected to 2D along the z-axis while performing a probabilistic update. A confusion matrix formulation is also applied (i.e. the error matrix associated with the semantic labels), to manage uncertainty and increase robustness.

The map representation is highly scalable and can be continuously updated over time. At the same time the generated semantic map has labels that can be understood by humans. It is a challenge for the vehicle to automatically determine a path based on the pixel-level semantic map, thus we explore its capability of being potentially used for navigation in this paper. 

\begin{figure*}
    \centering
        \includegraphics[scale=0.203]{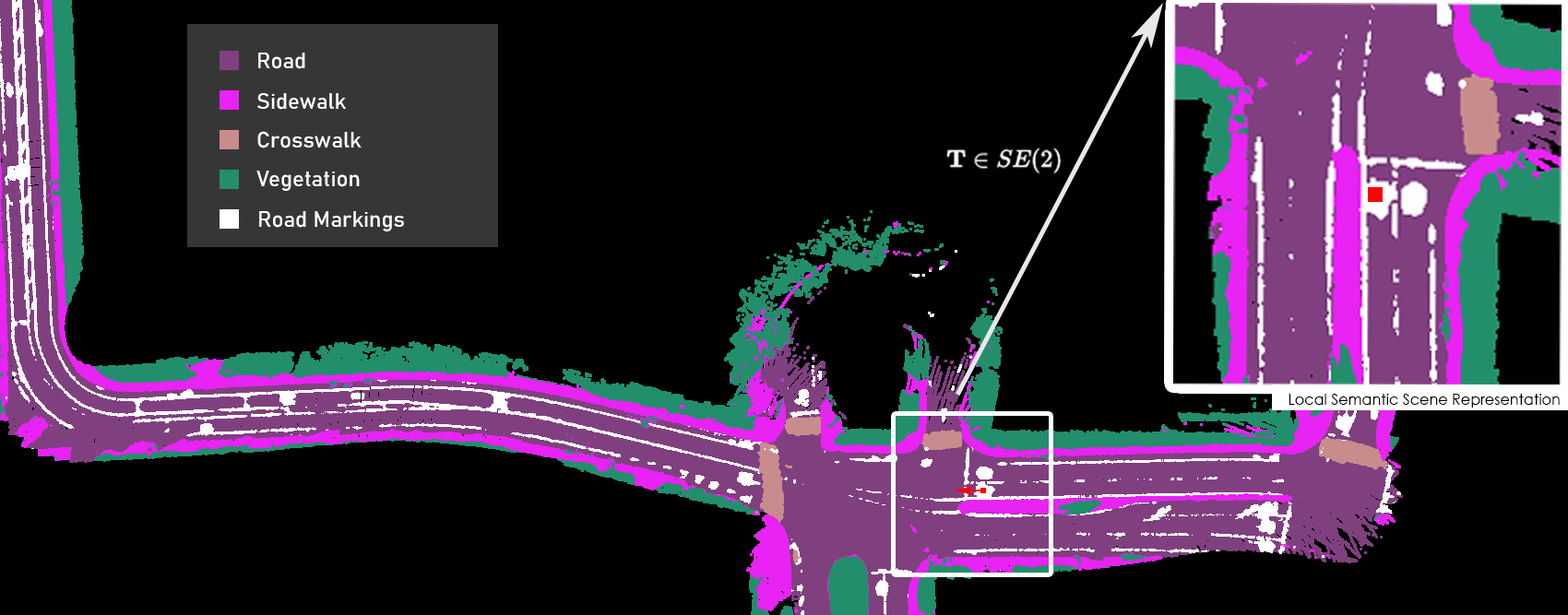}
    \caption{An example of the semantic map representation. An ego-centric representation is shown on the top right: red box denotes the rear-axle of the autonomous vehicle agent with an applied transformation $\mathbf{T}$.}
    \label{fig:semantic-model}
\end{figure*}

Once a semantic map has been generated, localization can be leveraged to estimate the ego-vehicle pose ($\mathbf{T} \in SE(2)$) and extract local semantic representations that account for the translation and orientation of the vehicle. An example is shown on Fig. \ref{fig:semantic-model}, where the vehicle is represented by the red square and the front of the car always faces upward on the image frame. 

\vspace*{-10mm}
\subsection{Conditional Variational Autoencoder}
\label{sec:cva}
As previously introduced in Section \ref{sec:related-forecast}, CVAEs have been used for estimating an external road user's future trajectory. One advantage of these models is that they allow us to explicitly encode the multi-modal aspects of trajectory prediction: based on a limited number of observations, a road user is likely to take multiple future trajectories. In other words, there is not a single best estimate in many cases. 

In this work, we extend CVAEs for dynamic trajectory generation. It is important to note that there are a few key differences between road user trajectory prediction and our approach. While trajectory generation is a multi-modal problem and the model must be capable of representing various modes for driving through intersections as an example, the intent and plan is known in advance and becomes deterministic when a high-level plan is given. In other words, the model should be capable of estimating the distribution of all possible modes and their corresponding trajectories but generate a single optimal trajectory given a high-level plan. 

In this context, we formulate the problem of identifying feasible trajectories $\mathbf{y}$ given a local scene representation and a global plan $\mathbf{m}$ as $p(\mathbf{y}\mid \mathbf{m})$.  
 
\begin{figure*}
    \centering
        \includegraphics[trim={0cm .15cm 0cm 0.2cm}, clip, scale=0.6]{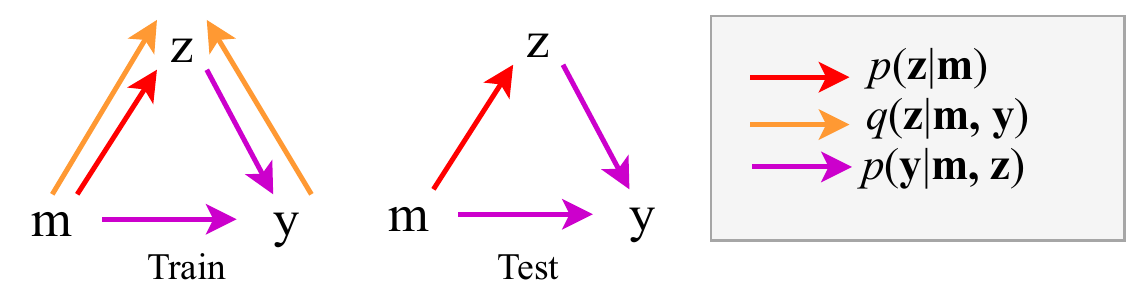}
    \caption{CVAE graphical model during training and testing.}
    \label{fig:graphical-model}
\end{figure*}

A CVAE is a directed graphical model (Fig. \ref{fig:graphical-model}) that is formulated using an input variable $\mathbf{m}$, an output variable $\mathbf{y}$, and an approximation for the distribution $p(\mathbf{y}\mid \mathbf{m})$.  To perform an approximation, a latent variable $\mathbf{z}$ is introduced and marginalized over all possible values as shown in (\ref{eq:1}), where $p(\mathbf{y}\mid \mathbf{m}, \mathbf{z})$ can be understood as a decoder and $p(\mathbf{z}\mid \mathbf{m})$ as an encoder. However, one problem arises while estimating $p(\mathbf{z}\mid \mathbf{m})$ as this requires inference of $p(\mathbf{m})$ as shown in (\ref{eq:2}).

To mitigate the intractable representation of the posterior $p(\mathbf{z}\mid \mathbf{m})$, a \textit{recognition} model $q(\mathbf{z}\mid \mathbf{m}, \mathbf{y})$ is introduced as an approximation that can be learned rather than treating it as a closed-form formulation. 

\begin{equation}
\label{eq:1}
p\left(\mathbf{y} \mid \mathbf{m}\right)=\sum_{\mathbf{z} \in \mathbf{Z}} p\left(\mathbf{y} \mid\mathbf{m}, \mathbf{z}\right) p\left(\mathbf{z} \mid \mathbf{m} \right)
\end{equation}

\begin{equation}
\label{eq:2}
p\left(\mathbf{z} \mid \mathbf{m}\right)=p\left(\mathbf{m} \mid \mathbf{z}\right) \cdot \frac{p\left(\mathbf{z}\right)}{p\left(\mathbf{m}\right)}
\end{equation}

Similar to \cite{leung} and \cite{ivanovic}, importance sampling can be performed by multiplying (\ref{eq:1}) with $q(\mathbf{z} \mid \mathbf{m}, \mathbf{y})/q(\mathbf{z} \mid \mathbf{m}, \mathbf{y})$ and rewriting the sum as an expectation over $q(\mathbf{z} \mid \mathbf{m}, \mathbf{y})$:

\begin{equation}
\label{eq:3}
\begin{aligned}
p(\mathbf{y} \mid \mathbf{m})=\mathbb{E}_{q}\left[\frac{p_{\phi}\left(\mathbf{y} \mid \mathbf{m}, \mathbf{z} \right) p_{\theta}(\mathbf{z} \mid \mathbf{m})}{q_{\psi}(\mathbf{z} \mid \mathbf{m}, \mathbf{y})}\right]
\end{aligned}
\end{equation}

For numerical stability reasons, during training the log-likelihood of $p(\mathbf{y} \mid \mathbf{m})$ is optimized as shown in (\ref{eq:4}) and due to the complexity of the right hand expression, Jensen's inequality is applied as shown in (\ref{eq:5}). Furthermore, the expression to be maximized can be rewritten using Kullback-Leibler divergence (\ref{eq:6}). The probability distributions are parameterized by different weights and are represented by subscripts $\psi$, $\phi$, and $\theta$.

\begin{equation}
\label{eq:4}
\log p(\mathbf{y} \mid \mathbf{m})=\log \mathbb{E}_{q}\left[\frac{p_{\phi}\left(\mathbf{y} \mid \mathbf{m}, \mathbf{z}\right) p_{\theta}(\mathbf{z} \mid \mathbf{m})}{q_{\psi}(\mathbf{z} \mid \mathbf{m}, \mathbf{y})}\right]
\end{equation}

\begin{equation}
\label{eq:5}
\log \mathbb{E}_{q}\left[\frac{p_{\phi}\left(\mathbf{y} \mid \mathbf{m}, \mathbf{z} \right) p(\mathbf{z} \mid \mathbf{m})}{q_{\psi}\left(\mathbf{z} \mid \mathbf{m}, \mathbf{y}\right)}\right] \geq \mathbb{E}_q \left[\log\frac{p_{\phi}\left(\mathbf{y} \mid \mathbf{m}, \mathbf{z} \right) p_{\theta}(\mathbf{z} \mid \mathbf{m})}{q_{\psi}(\mathbf{z} \mid \mathbf{m}, \mathbf{y})}\right]
\end{equation}

\begin{equation}
\label{eq:6}
\begin{aligned}
\log(\mathbf{y} \mid \mathbf{m}) \ge \mathbb{E}_{q}\left[\log p_{\phi}\left(\mathbf{y} \mid \mathbf{m}, \mathbf{z}\right)\right]- \mathbb{E}_{q}\left[\log q_{\psi}(\mathbf{z} \mid \mathbf{m}, \mathbf{y})-\log p_{\theta}(\mathbf{z} \mid \mathbf{m})\right]\\ 
= \mathbb{E}_{q}\left[\log p_{\phi}(\mathbf{y} \mid \mathbf{m}, \mathbf{z})]- \mathbb{KL}\left[q_{\psi}(\mathbf{z} \mid \mathbf{m}, \mathbf{y}) \| p_{\theta}(\mathbf{z} \mid \mathbf{m})\right]\right.
\end{aligned}
\end{equation}

While training, $p_{\theta}(\mathbf{z} \mid \mathbf{m})$, and $q_{\psi}(\mathbf{z} \mid \mathbf{m}, \mathbf{y})$ are jointly optimized and the latent variables $\mathbf{z}$ are sampled from $q_{\psi}(\mathbf{z} \mid \mathbf{m}, \mathbf{y})$. As previously derived in (\ref{eq:3})-(\ref{eq:6}), we seek to maximize (6) or in other words minimize $\left.\mathcal{L}_{\text {CVAE }}=-\mathbb{E}_{q} \log p_{\phi}\left(\mathbf{y} \mid \mathbf{m}, \mathbf{z}\right)\right]+\mathbb{KL}\left[q_{\psi}(\mathbf{z} \mid \mathbf{m}, \mathbf{y}) \| p_{\theta}(\mathbf{z} \mid \mathbf{m})\right]$. While this formulation alone can capture the multi-modal aspects of intersection navigation, we further improve the performance by incorporating a Mean-Squared-Error (MSE) loss term. Hence, the loss is given by:

\begin{equation}
\label{eq:7}
\left.\mathcal{L}=-\mathbb{E}_{q} [\log p_{\phi}\left(\mathbf{y} \mid \mathbf{m}, \mathbf{z}\right)\right]+\mathbb{KL}\left[q_{\psi}(\mathbf{z} \mid \mathbf{m}, \mathbf{y}) \| p_{\theta}(\mathbf{z} \mid \mathbf{m})\right]+\frac{1}{H} \sum_{i=1}^{H}\left\|\mathbf{y}_{i}-\hat{\mathbf{y}}_{i}\right\|^{2}
\end{equation}

On the other hand, during testing the model can estimate the complete distribution $p(\mathbf{y} \mid \mathbf{m})$ by explicitly computing (\ref{eq:1}) or the distribution $p(\mathbf{y} \mid \mathbf{m}, \mathbf{z}^*)$ that corresponds to the latent variable with the highest probability:

\begin{equation}
\label{eq:8}
\mathbf{z}^{*}=\arg \max _{\mathbf{z}} p_{\theta}(\mathbf{z} \mid \mathbf{m})
\end{equation}

Here $\mathbf{z}^*$ can be understood as the mode that best describes a deterministic trajectory. This is intuitive from the perspective of having various traversible trajectories for intersection navigation; however, by conditioning $\mathbf{z}$ on $\mathbf{m}$ -- the global plan and the local scene representation -- a single output is generated.

\subsection{Dynamic Path Generation using TridentNet}
\label{sec:dynamic-generation}
With the global plan, the local scene representation and the CVAE formulation in place, we focus on the implementation details for TridentNet. First, the global plan and the local scene representation are encoded using convolutional layers as shown in Fig. \ref{fig:tridentnet} . The dimensionality of these representations is further reduced using fully connected layers. 

Given that the \textit{recognition} model $q_{\psi}(\mathbf{z} \mid \mathbf{m}, \mathbf{y})$ is dependent on the ground truth trajectories ($\mathbf{y}$) during training, a bidirectional LSTM~\cite{lstm} is used to encode the next $H$ ground-truth positions. This embedding is then combined with the map representation embedding using another fully connected layer. The latent variable $\mathbf{z}$ that is sampled from $q_{\psi}(\mathbf{z} \mid \mathbf{m}, \mathbf{y})$ and $p_{\theta}(\mathbf{z} \mid \mathbf{m})$ is assumed to be drawn from a categorical distribution. 

Once $\mathbf{z}^*$ is determined following (\ref{eq:8}), $p(\mathbf{y} \mid \mathbf{m}, \mathbf{z}^*)$ is assumed to be a multivariate normal that is parameterized with a gated recurrent unit (GRU)~\cite{gru}: each multivariate normal is characterized by parameters ($\mathbf{\mu}_t$, $\mathbf{\Sigma}_t$) that are decoded from $t\rightarrow t+H$. Both graphical models that correspond to the training and testing process are shown in Fig. \ref{fig:graphical-model}.

\section{Data and Experiments}
\label{sec:experiments}
\begin{figure}[htb]
\begin{minipage}[t]{.4\textwidth}
\centering
    \includegraphics[width=\textwidth]{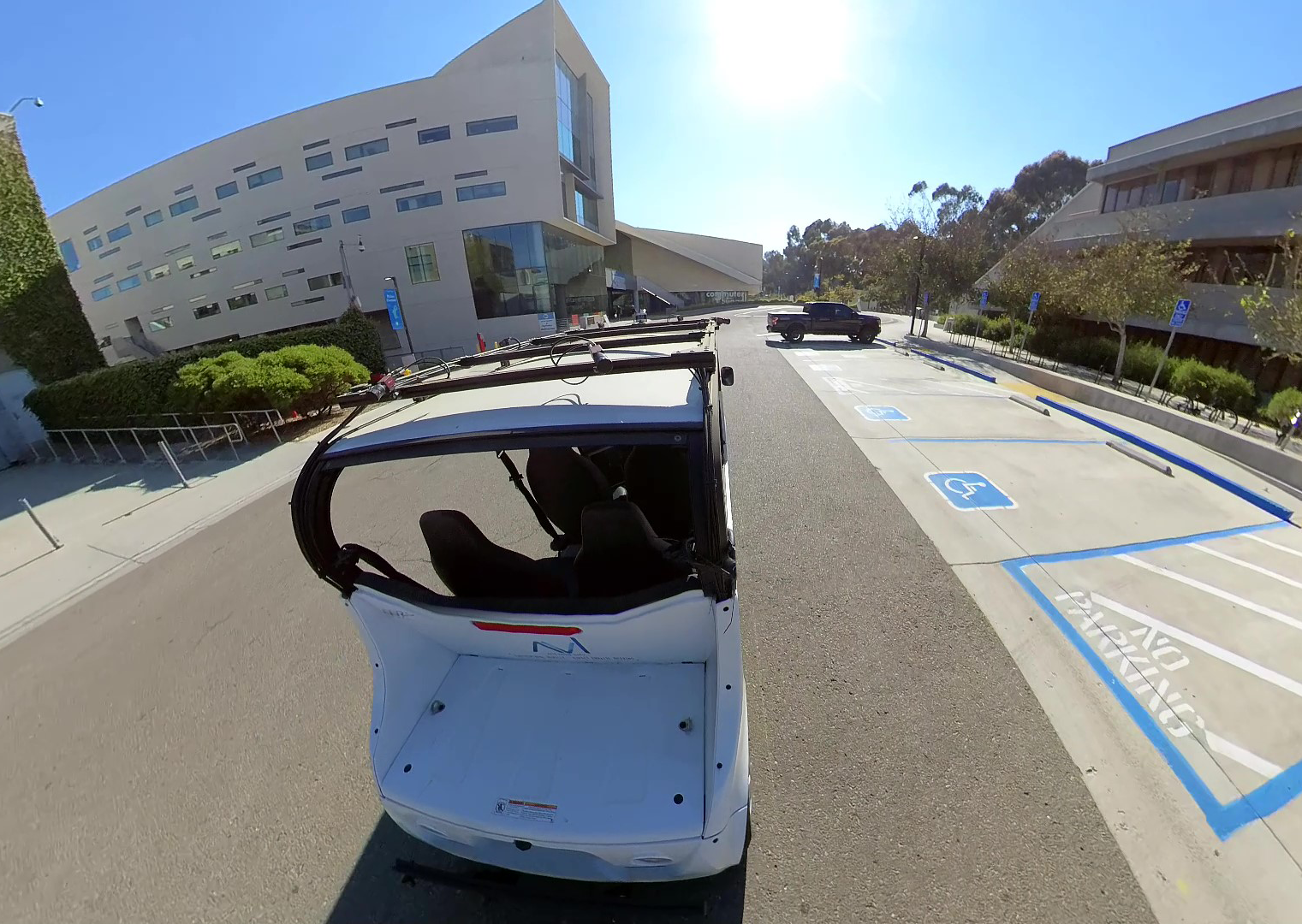}
    \subcaption{}
    \label{fig:data-collection}
\end{minipage}
\hfill
\begin{minipage}[t]{.4\textwidth}
    \centering
    \includegraphics[width=\textwidth]{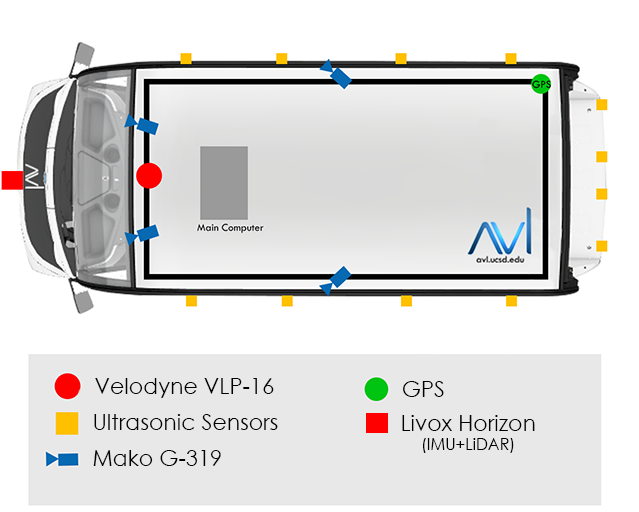}
    \subcaption{}
    \label{fig:vehicle-steup}
\end{minipage} 
\caption{Data collection process performed at UC San Diego using autonomous vehicle platform shown in (a). Its corresponding sensor configuration is shown in (b).}
\label{figure:2}
\end{figure}

To validate TridentNet, a data collection phase was performed at UC San Diego summer 2020 using a vehicle from UC San Diego's Autonomous Vehicle Laboratory. This data collection process consisted of three phases: calibration, mapping and data collection (Fig. \ref{fig:data-collection}). Our vehicle is equipped with various sensors as shown in Fig.~\ref{fig:vehicle-steup}; however, for this dataset, only the data from the front two cameras and a Velodyne VLP-16 were used for mapping, localization and scene modeling, and IMU, GPS, and Odometry data was used for generating the global plan representations. The semantic bird's eye view representations for three different parts of campus were generated following the procedures presented in \cite{paz:iros-semantic}. The data was annotated by leveraging localization; it is available online at avl.ucsd.edu and includes a training and test split.

\subsection{Calibration}
\label{sec:calibration}
As previously mentioned, one challenge with end-to-end models involves calibration and vehicle-specific assumptions. In order to address this, we leverage bird's eye view representations that can account for geometric transformations by first calibrating our sensors as described in~\cite{paz:iros-semantic}. All data for generating semantic maps was collected immediately after calibration. 

\subsection{Dataset and Ground Truth Labels}
\label{sec:dataset}
The dataset includes the global plans with routed/unrouted image representations, local semantic maps that have been extracted by localization, and the vehicle pose as a function of time. The global plan is represented with $200px \times 200px$ images with a $0.5m/px$ resolution. The local semantic scene model is represented using $400px \times 400px$ images with a $0.2m/px$ resolution and includes semantic class labels for road, crosswalk, sidewalk, lane markings and vegetation. The ground truth labels were automatically annotated by leveraging the localization of the vehicle and by manually driving through different parts of campus. Given that the vehicle pose is updated at approximately 10Hz, this can impact the distribution and the spacing of the ground truth labels. For this reason, we interpolate the data before training.

The training sequence includes a variety of different types of maneuvers including roads with relatively straight segments, high degree of curvature and even intersections and u-turns.

\subsection{Experiments}
\label{sec:experiments-experiments}
Given that our ground truth data can be interpolated by taking the future ego-
vehicle poses into account, we experiment with two different models: \textbf{H}10-\textbf{S}3 and \textbf{H}15-\textbf{S}2. \textbf{H}10-\textbf{S}3 corresponds to the interpolated ground truth ego-vehicle poses with a horizon of $10$ waypoints with an in-between waypoint distance of 3 meters; the training and test splits include $6,128$ and $2,864$ sequences, respectively. On the other hand, \textbf{H}15-\textbf{S}2 corresponds to a horizon of $15$ waypoints with a in-between waypoint distance of 2 meters; the training and test splits include $6,113$ and $2,860$ sequences, respectively. The number of modes $| \mathbf{z} |$ for the CVAE is chosen to be 12 for both models. Lower or higher values of $| \mathbf{z} |$ lead to slightly higher errors. While this parameter was not tuned extensively, the intuition behind this value is that it provides two modes per driving behavior: 1) intersection left-turns, 2) intersection right-turns, 3/4) driving straight with and without lane markings, 5) u-turns, and 6) curved roads.

\subsubsection{Quantifying Error}
\label{sec:errors}
To measure the performance of both models, we leverage the metrics from trajectory prediction literature \cite{metrics}: Average Displacement Error (ADE) and Final Displacement Error(FDE).\footnote{All units are given in meters.} These metrics measure the error between the generated waypoints and the ground truth waypoints: ADE (\ref{eq:9}) measures the average error across all of the waypoints generated over all test sequences. FDE (\ref{eq:11}) measures the average error for the last generated waypoint over all test sequences. Additionally, we measure the error for the first half of each sequence of waypoints separately (\ref{eq:10}) and the Maximum Displacement Error (\ref{eq:11}) to measure worst-case scenarios. The intuition for (\ref{eq:10}) is that during practical applications, a robot will perform path tracking on the initial waypoints immediately after generation and for that reason are more critical to safety.

\begin{equation}
\label{eq:9}
A D E_{\text{FULL}}=\frac{1}{n} \sum_{i=1}^{n} \frac{1}{H} \sum_{h=1}^{H}\left\|y_{i}^{h}-\hat{y}_{i}^{h}\right\|
\end{equation}

\begin{equation}
\label{eq:10}
A D E_{\text{HALF}}=\frac{1}{n} \sum_{i=1}^{n} \frac{1}{\left\lfloor{\frac{H-1}{2}}\right\rfloor} \sum_{h=1}^{\left \lfloor \frac{H-1}{2}\right\rfloor} \| y_{i}^{h}-\hat{y}_{i}^{h} \|
\end{equation}

\begin{equation}
\label{eq:11}
F D E=\frac{1}{n} \sum_{i=1}^{n}\left\|y_{i}^{H}-\hat{y}_{i}^{H}\right\|
\end{equation}

\begin{equation}
\label{eq:12}
M D E=\frac{1}{n} \sum_{i=1}^{n} \max_{h}\left\|y_{i}^{h}-\hat{y}_{i}^{h}\right\|
\end{equation}

The results for \textbf{H}10-\textbf{S}3 and \textbf{H}15-\textbf{S}2 are shown in Table \ref{table:1}. While both of the these models can generate a trajectory dynamically up to 30 meters,\footnote{Unlike trajectory forecasting literature, we associate the waypoints with an unit of distance instead of an unit of time due to the fact that the position of the waypoints over time will depend on the speed of the ego-vehicle. Furthermore, this allows us to decouple this module from downstream modules such as motion planning.} the GRU must decode additional waypoints for \textbf{H}15-\textbf{S}2 (15 waypoints) compared to \textbf{H}10-\textbf{S}3 (10 waypoints). It is observed that the model learns to generate smooth trajectories; however, for longer sequences (such as \textbf{H}15-\textbf{S}2) it is more likely to lead to compound errors as the next waypoint depends on the previous cell state. This leads to higher relative errors for \textbf{H}15-\textbf{S}2 as shown in Table \ref{table:1}. The highlighted values in the table indicate average errors in terms of meters: ADE$_{HALF}$ indicates an average of $0.3m$ per waypoint within the executable range for the ego-vehicle between $0m\rightarrow15m$. ADE$_{FULL}$ indicates an average of $1.1m$ per waypoint at a further range between $0m\rightarrow30m$. Additionally, we find that, in average, the worst case error (MDE) is closely related to the error of the final waypoint (FDE). In other words, the worst case error in the test set has a mode that corresponds to the last waypoint generated within the trajectory. 

\begin{figure}[htb]
\begin{minipage}[t]{.3\textwidth}
\centering
\includegraphics[trim={0 6cm 6cm 0}, clip, width=\textwidth]{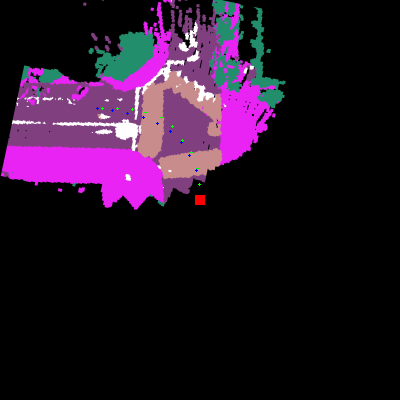}
\subcaption{}
\label{fig:H10-1}
\end{minipage}
\hfill
\begin{minipage}[t]{.3\textwidth}
\centering
\includegraphics[trim={2cm 4cm 4cm 2cm}, clip, width=\textwidth]{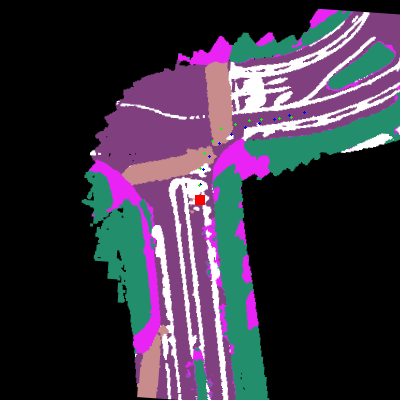}
\subcaption{}
\label{fig:H10-2}
\end{minipage}
\hfill
\begin{minipage}[t]{.3\textwidth}
\centering
\includegraphics[trim={2cm 4cm 4cm 2cm}, clip, width=\textwidth]{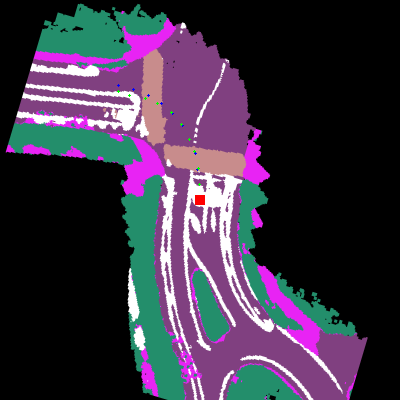}
\subcaption{}
\label{fig:H10-3}
\end{minipage}
\hfill

\caption{Testset Results for \textbf{H}10-\textbf{S}3. Ground truth waypoints are shown in blue; generated waypoints are shown in green. Images have been cropped for visualization; the rear-axle of the vehicle corresponds to the center of the red box.}
\label{figure:test-vis}
\end{figure}

While verification was not performed on the autonomous agent, these low relative errors indicate feasible and executable trajectories that can be visually verified. Furthermore, to showcase the multi-modal capabilities of our model, three intersection navigation test data samples generated by \textbf{H}10-\textbf{S}3 are shown in Fig. \ref{figure:test-vis}. In the figure, the green waypoints are dynamically generated by TridentNet whereas the blue waypoints are ground truth interpolated waypoints. Both trajectories are first expressed with respect to the rear-axle of the vehicle in meters then reprojected onto the local semantic representation using a $0.2m/px$ discritization factor.

\begin{table}
\begin{tabular}{ |p{3cm}||p{2.3cm}|p{2.3cm}|p{2cm}|p{2cm}|  }
 \hline
 \multicolumn{5}{|c|}{UC San Diego Dataset} \\
 \hline
 Model & ADE$_{FULL}(m)$ & ADE$_{HALF}(m)$ & FDE(m) & MDE(m) \\
 \hline
 TridentNet-\textbf{H}10-\textbf{S}3   & $\mathbf{1.056245}$  &   $\mathbf{0.336941}$  &  $\mathbf{2.447714}$ & $\mathbf{2.494614}$\\
 TridentNet-\textbf{H}15-\textbf{S}2   & 2.341875  & 0.753802    &   6.127183 & $6.177646$\\
 \hline
\end{tabular}
 \caption{Evaluation results for Trajectory Generation on test set. Waypoint units are given in meters.}
 \label{table:1}
\end{table}

\vspace*{-15mm}
\section{Conclusion and Future Work}
\label{sec:summary}
We introduced an approach that leverages a conditional generative model for dynamically generating feasible trajectories for autonomous driving. Unlike current state of the art architectures for automated driving, our approach does not rely on route-network definitions annotated at a centimeter-level (i.e. HD maps). Instead, our approach leverages lightweight semantic maps that can be continuously updated automatically. Furthermore, unlike end-to-end deep learning models, our approach accounts for arbitrary sensor configurations, operates using a single nominal representation instead of raw image data to represent the scene, and does not assume vehicle specific control inputs. Our experiments with the collected data show that the proposed model can generate trajectories with low relative error.

For future work, we plan to design an open-source hierarchical framework that leverages TridentNet and conduct live experiments on full-scale autonomous vehicle platforms to evaluate the real-time capabilities of the system. Given that our proposed method is decoupled from vehicle specific configurations and provides a clear interface for perception and motion planning, our work can enable additional experimentation with road user trajectory prediction (forecasting) and obstacle avoidance across various types of autonomous agents. While our initial design strategy for TridentNet utilizes a semantic map to model the scene and provide contextual information, we plan to experiment with alternative scene representations that can be estimated in real-time. 

%
%
%
%
%


\begin{thebibliography}{5}
%
\bibitem{nuscenes}
Caesar, H., Bankiti, V., Lang, A. H., Vora, S., Liong, V. E., Xu, Q., Krishnan, A., Pan, Y., Baldan, G., and Beijbom, O. (2020). NuScenes: A multimodal dataset for autonomous driving. In Proceedings of the IEEE/CVF Conference on Computer Vision and Pattern Recognition (pp. 11621-11631).

\bibitem{argoverse}
Chang, M. F., Lambert, J., Sangkloy, P., Singh, J., Bak, S., Hartnett, A., Wang, D., Carr, P., Lucey, S., Ramanan, D., and Hays, J. (2019). Argoverse: 3d tracking and forecasting with rich maps. In Proceedings of the IEEE Conference on Computer Vision and Pattern Recognition (pp. 8748-8757).

\bibitem{apollo}
ApolloAuto/apollo. GitHub. (2020). Retrieved 20 December 2020, from https://github.com/ApolloAuto/apollo.

\bibitem{autoware}
Kato, S., Takeuchi, E., Ishiguro, Y., Ninomiya, Y., Takeda, K. and Hamada, T. (2015). "An Open Approach to Autonomous Vehicles," in IEEE Micro, (vol. 35, no. 6, pp. 60-68).

\bibitem{hecker:one}
Hecker, S., Dai, D., and Van Gool, L. (2018). End-to-end learning of driving models with surround-view cameras and route planners. In Proceedings of the European Conference on Computer Vision (ECCV) (pp. 435-453).

\bibitem{hecker:two}
Hecker, S., Dai, D., Liniger, A., Hahner, M., and Van Gool, L. (2020). Learning Accurate and Human-Like Driving using Semantic Maps and Attention. IEEE/RSJ International Conference on Intelligent Robots and Systems (IROS) (pp. 2346-2353).

\bibitem{rus:one}
Amini, A., Rosman, G., Karaman, S., and Rus, D. (2019). Variational end-to-end navigation and localization. In 2019 International Conference on Robotics and Automation (ICRA) (pp. 8958-8964). 

\bibitem{rus:two}
Amini, A., Gilitschenski, I., Phillips, J., Moseyko, J., Banerjee, R., Karaman, S., and Rus, D. (2020). Learning Robust Control Policies for End-to-End Autonomous Driving from Data-Driven Simulation. IEEE Robotics and Automation Letters (pp. 1143-1150).

\bibitem{capaso:iv}
Capasso, A. P., Bacchiani, G., and Broggi, A. (2020). From Simulation to Real World Maneuver Execution using Deep Reinforcement Learning. IEEE Intelligent Vehicles Symposium (IV), Las Vegas, NV, USA, (pp. 1570-1575).

\bibitem{can}
Can, Y. B., Liniger, A., Unal, O., Paudel, D., and Van Gool, L. (2020). Understanding Bird's-Eye View Semantic HD-Maps Using an Onboard Monocular Camera. arXiv preprint arXiv:2012.03040.

\bibitem{roddick}
Roddick, T., and Cipolla, R. (2020). Predicting Semantic Map Representations from Images using Pyramid Occupancy Networks. In Proceedings of the IEEE/CVF Conference on Computer Vision and Pattern Recognition (pp. 11138-11147).

\bibitem{deeplab}
Chen, L. C., Zhu, Y., Papandreou, G., Schroff, F., and Adam, H. (2018). Encoder-decoder with atrous separable convolution for semantic image segmentation. In Proceedings of the European conference on computer vision (ECCV) (pp. 801-818).

\bibitem{paz:hd-maps}
Paz Ruiz, D. (2020). Autonomous Vehicles: Their Capabilities and Limitations. Diss. UC San Diego.

\bibitem{paz:iros-semantic}
Paz, D., Zhang, H., Li, Q., Xiang, H., and Christensen, H. (2020). Probabilistic Semantic Mapping for Urban Autonomous Driving Applications. In International Conference on Intelligent Robots and Systems (IROS), IEEE/RSJ.

\bibitem{paz:fsr}
Paz, D., Lai, P. J., Harish, S., Zhang, H., Chan, N., Hu, C., Binnani, S., and Christensen, H. (2019). Lessons learned from deploying autonomous vehicles at UC San Diego. Field and Service Robotics.

\bibitem{alahi}
Alahi, A., Kratarth, G., Ramanathan, V., Robicquet A., Li, F., Savarese, S. (2016). Social LSTM: Human Trajectory Prediction in Crowded Spaces. Proceedings of the IEEE Conference on Computer Vision and Pattern Recognition (CVPR) (pp. 961-971).

\bibitem{leung}
Ivanovic, B., Leung, K., Schmerling, E., and Pavone, M. (2020). Multimodal Deep Generative Models for Trajectory Prediction: A Conditional Variational Autoencoder Approach, IEEE Robotics and Automation Letters (In Press).

\bibitem{ivanovic}
Salzmann, T., Ivanovic, B., Chakravarty, P., and Pavone, M. (2020). Trajectron++: Dynamically-feasible trajectory forecasting with heterogeneous data. arXiv preprint arXiv:2001.03093.

\bibitem{lstm}
Hochreiter, S., Schmidhuber, J. (1997). Long short-term memory. Neural Computation.

\bibitem{gru}
Cho, K., van Merrienboer, B., Gulcehre, C., Bahdanau, D., Bougares, F., Schwenk, H., Bengio, Y. (2014). Learning phrase representations using rnn encoder-decoder for statistical machine translation. Proceedings of the 2014 Conference on Empirical Methods in Natural Language Processing. (pp. 1724–1734).

\bibitem{gupta}
Gupta, A., Johnson, J., Fei-Fei, L., Savarese, S., and Alahi, A. (2018). Social gan: Socially acceptable trajectories with generative adversarial networks. In Proceedings of the IEEE Conference on Computer Vision and Pattern Recognition (pp. 2255-2264).

\bibitem{yu}
Yu, B., Yin, H., and Zhu, Z. (2017). Spatio-temporal graph convolutional networks: A deep learning framework for traffic forecasting. arXiv preprint arXiv:1709.04875.

\bibitem{yu2}
Walters, R., Li, J., and Yu, R. (2021). Trajectory Prediction using Equivariant Continuous Convolution. International Conference on Learning Representations (ICLR).




\bibitem{metrics}
Yang, T., Nan, Z., Zhang, H., Chen, S., and Zheng, N. (2020). Traffic Agent Trajectory Prediction Using Social Convolution and Attention Mechanism. arXiv preprint arXiv:2007.02515.



\end{thebibliography}
\end{document}